\pdfoutput=1

\documentclass[11pt]{article}
\usepackage{multirow}
\usepackage[table,xcdraw]{xcolor}
\usepackage{graphicx}
\usepackage{enumitem}
\usepackage{footnote}
\makesavenoteenv{tabular}
\usepackage{amsmath}
\usepackage{hyperref}
\usepackage{float}
\usepackage{booktabs}
\usepackage{tabularx}
\usepackage[normalem]{ulem}
\usepackage[final]{acl}

\usepackage{times}
\usepackage{latexsym}

\usepackage[T1]{fontenc}

\usepackage[utf8]{inputenc}

\usepackage{microtype}

\usepackage{inconsolata}

\usepackage{graphicx}

\title{Learning from \textit{Sufficient} Rationales: Analysing the Relationship Between Explanation Faithfulness and Token-level Regularisation Strategies}

\author{
    \textbf{Jonathan Kamp\textsuperscript{1}} \hspace{1cm}
    \textbf{Lisa Beinborn\textsuperscript{2}} \hspace{1cm}
    \textbf{Antske Fokkens\textsuperscript{1}}
    \\
    \textsuperscript{1}Computational Linguistics and Text Mining Lab, Vrije Universiteit Amsterdam\\
    \textsuperscript{2}Institute of Computer Science, University of Göttingen\\
    \texttt{\{j.b.kamp, antske.fokkens\}@vu.nl, lisa.beinborn@uni-goettingen.de}
    }

\begin{document}
\maketitle
\begin{abstract}
Human explanations of natural language, \textit{rationales}, form a tool to assess whether models learn a label \textit{for the right reasons} or rely on dataset-specific shortcuts. \textit{Sufficiency} is a common metric for estimating the informativeness of rationales, but it provides limited insight into the effects of rationale information on model performance. We address this limitation by relating sufficiency to two modelling paradigms: the ability of models to identify which tokens are part of the rationale (through token classification) and the ability of improving model performance by incorporating rationales in the input (through attention regularisation). 
We find that highly informative rationales are not likely to help classify the instance correctly. Sufficiency conversely captures the classification impact of the non-rationalised context, which interferes with rationale information in the same input.
We also find that incorporating rationale information in model inputs can boost cross-domain classification, but results are inconsistent per task and model type. Finally, sufficiency and token classification appear to be unrelated. These results exemplify the complexity of rationales, showing that metrics capable of systematically capturing this type of information merit further investigation.


\end{abstract}

\section{Introduction}

Neural text classifiers are trained to generalise patterns in texts to predict labels for unseen examples. The learnt patterns optimise the classification objective but do not necessarily align with a human interpretation of the task \citep{jakobsen2023being, ross2017right}. For example, an argument classification model tends to overfit to topic-specific vocabulary instead of capturing general argumentation structure. To uncover such over-simplified classification patterns, we can compare model behaviour to human-annotated reasons for input labels, i.e.\ \textit{rationales} \citep{carton-etal-2020-evaluating, strout2019human}. When provided in \textit{highlighted} form \citep{wiegreffe-marasovic-2021-review} rather than e.g.\ free-text, token-level rationales establish a benchmark for analysing model behaviour at high granularity patterns within the input, such as words and phrases. 

\begin{figure}[tbp]
    \centering
    \includegraphics[width=0.48\textwidth]{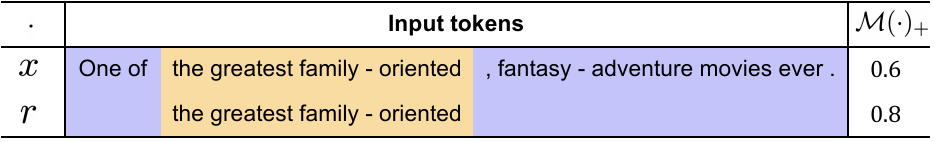}
    \caption{We compare the model $\mathcal{M}$ probability of the gold label $+$ when using the full input $x$ to the probability on the isolated rationales $r$. The example is an instance from the sentiment analysis dataset SST \citep{socher-etal-2013-recursive, carton-etal-2020-evaluating}.}
    \label{fig:figure1}
\end{figure}



\textit{Sufficiency} \citep{lei2016rationalizing} is commonly adopted to quantify the contribution of highlight rationales to model predictions, i.e., \textit{rationale informativeness}, but the interpretation of this metric remains opaque. In principle, sufficiency corresponds to the change in model confidence when reducing the input to the rationale. Its use is prevalent as a \textit{faithfulness} proxy for post-hoc attribution methods \citep{manna-sett-2024-faithfulness, kamahi-yaghoobzadeh-2024-counterfactuals}, automated techniques that approximate each input token's informativeness to the model's prediction. By assessing their informativeness, sufficiency measures how accurately these artificial token-level rationales truly reflect the model's inner workings.
In Figure~\ref{fig:figure1}, the probability for positive sentiment is higher for rationales $r$ ("the greatest family-oriented") compared to input $x$, 
suggesting that, in this case, the non-rationale words constitute noise that steers the probability away from positive sentiment.
However, the current use of sufficiency is limited to a \textit{relative} faithfulness estimation when ranking attribution methods, while \textit{high} and \textit{low} sufficiency scores remain vague in the absolute sense. In fact, such use (i) oversimplifies the interaction between important and (seemingly less relevant) contextual linguistic cues in the input, and (ii) does not provide guidance into how rationale informativeness can be exploited by a model; we address this limitation.

We investigate the role that sufficiency plays in capturing rationale informativeness by establishing its relation to two modelling paradigms with rationale inputs. In particular, we investigate the relation between sufficiency and (a) the ability of a model to identify tokens that are part of the rationale and (b) the ability of rationales to improve a model's performance. The way in which these paradigms profit from rationales is expected to reflect how sufficiency captures rationale informativeness. Although input rationalisation has been shown to improve both in- and cross-domain classification \citep{hartmann-sonntag-2022-survey}, under which conditions rationales are helpful remains an open question \citep{hase2022can}. In addition to exploring the relation to sufficiency, this paper is, to our knowledge, also first to systematically assess the effect of adding rationale information to input across tasks, models and learning strategies. We have mixed results, though our \texttt{BERT} model benefits from rationales in a cross-domain setting.

Our results do not confirm the hypothesis that rationales that are highly informative (based on their sufficiency score) consist of easily identifiable tokens, nor the hypothesis that good sufficiency scores predict that rationales improve classification. Instead, our research reveals that sufficiency captures the impact of non-rationale contexts on model predictions. This highlights that the relation between rationales and their context is complex and that sufficiency by itself can only capture relative information.\footnote{We publicly release our code at\\ 
\url{https://github.com/jbkamp/repo-Suff-Rationales}.}




\section{Related Work}
We first describe sufficiency's role as a key metric for model understanding (§\ref{sec:faithfulness_assessment}) and then describe how rationales have been used to improve model performance in previous work (§\ref{sec:learning_from_rationales}).

\subsection{Sufficiency for Faithfulness Assessment}\label{sec:faithfulness_assessment}
Token-level explanations are subject to inter-rater variability, both when the rater is human or artificial. When explanations are extracted artificially from a fine-tuned model, e.g. via saliency methods, assessing their \textit{faithfulness} to the inner decision making processes is crucial to model interpretability. This is especially relevant given the tendency of explanations to disagree, even on the same instance \citep{neely2022song,kamp-etal-2024-role}. Research on faithfulness assessment keeps being relevant; for example, \citet{fayyaz2024evaluating} show the importance of faithfulness evaluation for Large Language Models (LLMs), finding that feature attributions are more faithful than prompt-based self-explanations and align better with rationales. \citet{Madsen-etal-2024-self} propose consistency checks for self-explanations. A question regarding human explanations, instead, is whether they are informative enough for a model to learn from. \textit{Sufficiency} \citep{lei2016rationalizing} serves as a key metric for both human and model explanations.


In line with other faithfulness metrics, sufficiency is ablation-based. In \citet{deyoung2020eraser}, collected rationales are assessed on sufficiency based on the notion of contrast examples \citep{zaidan2007using}: rationales are isolated from their context, resulting in a change of probability towards the target class of a test instance. Conversely, \textit{comprehensiveness} \citep{yu2019rethinking} measures the change in probability by removing the rationales. Similarly, \citet{hooker2019benchmark} observe the effect of removing important features by training different models. We investigate sufficiency (formalised in §\ref{sec:sufficiency}) as it may indirectly estimate the importance of both rationale words and context words.

Calculating sufficiency may lead to out-of-distribution examples when non-rationales are removed or masked \citep{madsen2024-faithfulness-measurable, carton2022learn, hooker2019benchmark}. In fact, when we do so, we often end up with incomplete and ungrammatical sentences causing irregular model behaviour. A handful of studies investigate the difference between removing and masking non-rationales on a single model type \citep{kamahi-yaghoobzadeh-2024-counterfactuals, carton2022learn}. We find no differences for models with different pre-training objectives.


The unique contribution of our work is that we explicitly link sufficiency to model improvement, investigating both masked language models and autoregressive decoders. We bridge the gap from rationale characteristics, which we specify as informativeness to a prediction, to learnability.

\subsection{Learning from Rationales}\label{sec:learning_from_rationales}



The power of deep learning lies in learning patterns in the data without explicit instructions. A commonly undesired consequence is the tendency for models to learn dataset-specific shortcuts. Human rationales can be used as an additional source of task-specific information to steer the learning process. \citet{hase2022can} compare different types of rationales and find that rationales as highlights are more beneficial for improving fine-tuned performance of a retrieval task than free-text explanations. They experiment with a synthetic task showing that retrieval-based RoBERTa models \citep{liu2019roberta, reimers-gurevych-2019-sentence} improve with explanations encoded in the input. Recent work on learning from rationales has been implemented on BERT-like models through attention regularisation. The classifier jointly computes the main loss term and a second term optimising the attention weights to a given input rationale mask, often provided by feature attribution methods. 
Results from previous studies are heterogeneous. Some show that additional rationale information is beneficial to the task: for example, \citet{attanasio-etal-2022-entropy} find that rationalisation mitigates unintended gender bias in hate speech detection and favours cross-domain generalisation; \citet{stacey2022supervising} find improvements on both in- and cross-domain natural language inference; \citet{carton2022learn} find accuracy improvements between 0.4\% and 3\% on three different in-domain tasks; \citet{krishna2023post} and \citet{bhan-etal-2024-self} find that attribution methods' explanations (fed back as additional prompt context) can improve instruction-tuned LLMs such as Mistral-7B \citep{jiang2023mistral7b}. In contrast, \citet{ferreira-etal-2025-explanation} find that guiding the model with rationales does not necessarily improve on an out-of-domain classification task; \citet{kozlova2024transformer} exploit eyetracking data to guide encoder-only and encoder-decoder models on an anaphora resolution task, finding mostly no improvement on the baseline models. 


In a similar fashion, \citet{su2024codeart} adopt attention regularisation in pre-training to guide a transformer when crucial cues (comments, function names) to code understanding are absent. In argument mining, \citet{jayaram-allaway-2021-human} fine-tune a stance predictor and use rationales as priors to optimise MAW, a feature attribution method based on the model's attention weights, in producing more human-like explanations. While they show that MAW is relatively \textit{faithful} compared to Gradient\,×\,Input \citep{shrikumar2016not} and a random baseline, the link between rationale faithfulness and informativeness remains mostly unclear. Comparably, \citet{xie-etal-2024-ivra} fine-tune Electra \citep{clark2020electra}, Llama-2-7b \citep{touvron2023llama}, and GPT-2-medium \citep{radford2019language} to produce explanations by regularising the attention distribution via different loss terms that optimise for faithfulness criteria (such as \textit{sufficiency}). They outperform other types of model guidance. This shows that sufficiency can be a beneficial component for learning through attention regularisation, which we will explore further.

\citet{hartmann-sonntag-2022-survey} train a token classifier simultaneously with a regular classifier (in a multi-task learning fashion) to produce explanations for the main task. \citet{ross-etal-2022-self} find that this technique, also called \textit{self-rationalisation}, produces task- and model-specific effects. \citet{pruthi-etal-2022-evaluating} find that attention regularisation outperforms the multi-task learning approach; this suggests that either the token classifier is inadequate to learn enough from the input rationales, or the learnt information is not fully utilised through the shared parameters of the main task. \citet{carton-etal-2020-evaluating} evaluate rationale quality; however, the link between rationale characteristics and the ability of a token classifier to distinguish rationales from non-rationales in a sequence is currently underdefined. To our knowledge, different approaches that use rationales for model improvement have not yet been systematically compared for a wider range of tasks.

\section{Data and Models}\label{sec:experimental_setup}

We describe the classification tasks and model types for which we compute sufficiency (expected to quantify rationale information) and for which we assess the ability to learn from rationales. 
By combining these two aspects in later sections, we aim to understand the way in which sufficiency captures rationale informativeness. 

\subsection{Data}\label{sec:datasets}

The inputs for our models are \textit{rationalised inputs}: tokenised texts with an accompanying binary mask indicating whether a token is part (1) or not part (0) of the rationale. Sets of rationalised inputs can be created in two ways. (i) Rationales are added to labeled sentences (or larger units), e.g. e-SNLI \citep{camburu2018snli} and HateXplain \citep{mathew2021hatexplain}. (ii) Token-level labels are repurposed \citep{wiegreffe-marasovic-2021-review} such as in \mbox{AURC-8} \citep{trautmann2020fine}, where sequences of tokens labeled as argument units from the original dataset become the rationales for the new task of predicting whether a sentence contains an argument. We aim for a representative sample of tasks selecting four existing datasets and adding two new repurposed ones. They are presented in Table~\ref{tab:data_charact} and described below.



\begin{table}[ht]
\small
\centering
\begin{tabularx}{\columnwidth}{l *{4}{>{\centering\arraybackslash}X}}
\toprule
\textbf{Dataset} & \textbf{\#cl} & \textbf{gran} & \textbf{type} & $|r|/|x|$ \\
\midrule
AURC-8      & 2 & None   & Semantic  & .41 \\
SST         & 2 & None   & Semantic  & .46 \\
HateXplain  & 3 & Phrase & Semantic  & .19 \\
e-SNLI      & 3 & Word   & Both & .21 \\
CoNLL-chunk  & 2 & Phrase & Syntactic  & .01 \\
CoNLL-NER  & 2 & None   & Syntactic  & .07 \\
\bottomrule
\end{tabularx}
\caption{The datasets vary in: \textbf{\#cl}asses (2 or 3); the \textbf{gran}ularity restriction in the annotations (none, word- and phrase-level, following \citet{wiegreffe-marasovic-2021-review}); the \textbf{type} of linguistic cues the model is expected to mostly rely on (syntactic or semantic, following \citet{jang-etal-2024-study-attention}); the rationale density: the average number of rationale tokens over input tokens ($|r|/|x|$).}
\label{tab:data_charact}
\end{table}

\paragraph{AURC-8} The texts from \citet{trautmann2020fine} are labelled as \textit{argumentative} or \textit{non-argumentative} in the context of a given topic or \textit{domain} (e.g. minimum wage). We explore both original topic-based splits of the data: in-domain (AURC-8\textsubscript{\textsc{id}}; 4,193 training and 1,200 test instances on same topics) and cross-domain (\mbox{AURC-8\textsubscript{\textsc{cd}}}; 4,000 training and 2,000 test instances on different topics). 

\paragraph{Stanford Sentiment Treebank (SST)} Following \citet{carton-etal-2020-evaluating}, we obtain the rationales from SST \citep{socher-etal-2013-recursive} by flattening the original syntactic phrase-level annotations of movie review snippets (see Appendix \ref{appendix:rationale_derivation_algorithm_SST} for details). The labels are positive and negative. There are 6,917 training and 1,821 test instances.


\paragraph{HateXplain} Texts in this set are labelled as \textit{hatespeech}, \textit{offensive}, \textit{normal} \citep{mathew2021hatexplain}. For instances with multiple annotators, we aggregate the labels by majority vote (in line with the authors) and the rationales by taking the union ($\cup$) or intersection ($\cap$). 
We aggregate rationale annotations to obtain binary rationale masks over input tokens, in line with the other datasets that we cover.
HateXplain\textsubscript{$\cup$} allows for human variation, while HateXplain\textsubscript{$\cap$} restricts the rationale to the tokens that were highlighted by all annotators.  Each version has 15,379 training and 1,924 test instances.

\paragraph{e-SNLI} This dataset by \citet{camburu2018snli} contains highlight rationales to explain a natural language inference task. The labels (entailment, contradiction, neutral) indicate the semantic relationship between two sentences that are concatenated by a [SEP] token. It contains 549,339 training and 9,824 test instances.

\paragraph{CoNLL-chunk} We re-purpose the CoNLL-2000 chunking dataset \citep{tjong-kim-sang-buchholz-2000-introduction}. For each type of syntactic phrase, we create a version of the dataset where the task is to predict its presence or absence. We ignore the three most frequent tags (\textsc{np, vp, pp}) as they would lead to barely any 0-class sentence labels. We choose the next-most frequent tags with at least 500 occurrences: \textsc{advp, adjp, sbar, prt}. Each version (same texts, different labels) has 8,936 training and 2,012 test instances.

\paragraph{CoNLL-NER} We re-purpose the CoNLL-2003 NER dataset \citep{tjong-kim-sang-de-meulder-2003-introduction}. For each of the tags \textit{person} (\textsc{per}), \textit{organisation} (\textsc{org}), \textit{location} (\textsc{loc}), \textit{miscellaneous} (\textsc{misc}), we create a version of the dataset where the task is to predict its presence or absence. Each has 14,041 training and 3,453 test instances.

\subsection{Models}\label{sec:models}

For each classification task, we fine-tune four transformer models that have a comparable number of parameters: \textbf{\texttt{BERT}}--base-uncased, 110M \citep{devlin-etal-2019-bert}, \textbf{\texttt{Pythia}}--160M \citep{biderman2023pythia}, \textbf{\texttt{ModernBERT}}--base, 149M \citep{modernbert}, \textbf{\texttt{GPT-Neo}}--125M \citep{gpt-neo}. Each is fine-tuned three times per task with three different random seeds. Similarly, \citet{wyatte2024scaling} compare bert-large, 330M to Pythia, 410M, which also fall into the same sizes. Among the 70M and 160M Pythia we opt for the latter, being the more popular\footnote{146k vs. 120k downloads until December 2024.} middle-sized model of the two. The runtimes of the four models fall in comparable ranges (Appendix \ref{appendix:runtimes_and_hardware}). \texttt{BERT} and \texttt{ModernBERT} use bidirectional attention and are therefore expected to perform better than our left-to-right attention models on the classification tasks, although mixed findings emerge from previous work \citep{bouchiha2025hierarchical, pilicita2025llms, lukito-etal-2024-comparing}.

\section{Sufficiency as Contextual Impact}\label{sec:sufficiency}
To better understand the added value of rationale information and the role of sufficiency as rationale informativeness proxy, we first formalise sufficiency and compute a corresponding value for each instance in the different datasets. We calculate sufficiency as the confidence change of the model $\mathcal{M}$, by measuring the change in probability between the prediction on instance $x_i$ and its counterexample $r_i$ for a given class $j$, in line with \citet{deyoung2020eraser} (see example in Figure \ref{fig:figure1}):
\begin{equation}\label{eq:sufficiency_equation}
    \text{suff}(x_i) = \mathcal{M}(x_i)_j - \mathcal{M}(r_i)_j
\end{equation}

\noindent We average sufficiency over all instances in a dataset ($D$) for a specific model $\mathcal{M}$:
\begin{equation}
    \text{suff}(D) = \frac{1}{n} \sum_{i=1}^{n} \left( \mathcal{M}(x_i)_j - \mathcal{M}(r_i)_j \right)
\end{equation}

\noindent Here, $r_i$ corresponds to $x_i$ in which the non-rationale context tokens are ablated. For context ablation, we apply removal or masking; we mask by using the pre-trained [MASK] token for \texttt{BERT} and \texttt{ModernBERT} and a random embedding for \texttt{Pythia} and \texttt{GPT-Neo} (details in Appendix~\ref{appendix:technical_details}). As the two implementations produce similar scores (Appendix~\ref{appendix:secondary_results}), we will only cover removal in our analyses.
Sufficiency has been defined such that a value of zero indicates that the rationale alone is strictly sufficient to predict the correct label. Higher values indicate that the other context tokens also contribute important information to the prediction. We find it more intuitive to interpret the metric as an indicator for \textit{\textbf{contextual impact}} ($CI$), i.e.,\ if the value approaches zero, the context does not have additional impact on the prediction (compared to the rationale). High $CI$ is expected to entail low rationale informativeness.

\begin{figure}[htbp]
    \centering
    \includegraphics[width=0.49\textwidth]{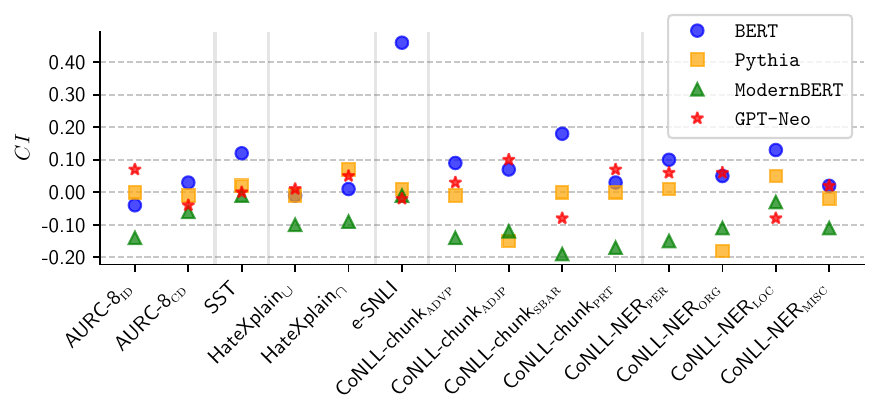}
    \caption{Dataset-average contextual impact ($CI$).}
    \label{fig:raw_sufficiency}
\end{figure}

Figure \ref{fig:raw_sufficiency} provides an overview of the dataset-average $CI$ for different tasks and models. For example, the large difference between $CI$ for \texttt{BERT} and the other models on e-SNLI may show that different models encode the balance between context and rationale information differently even on the same data. The overall positive $CI$ by \texttt{BERT} across tasks might indicate that the model relies more on context information than e.g.\ \texttt{ModernBERT}, a model characterised by an overall negative $CI$. This difference might stem from a different model behaviour on missing context (e.g.\ linguistic connectors) when predicting $\mathcal{M}(r_i)_j$ in Equation \ref{eq:sufficiency_equation}, where $r_i$ is essentially a bag-of-rationales with preserved order from the source sentence. In §\ref{sec:modelling_with_rationales}, we investigate to what extent $CI$ captures rationale informativeness for learning these tasks.


\section{Learning from Rationale Information}\label{sec:modelling_with_rationales}


We investigate the model's ability to learn from rationales on the six datasets using two learning paradigms: \uline{t}oken \uline{c}lassification (with metric $TC$) and \uline{a}ttention \uline{r}egularisation (with metric $AR$). The first paradigm investigates the ability of models to distinguish rationale tokens from the adjacent context. The second paradigm investigates whether including information from rationales in the input improves the model's performance.


\subsection{Learnability Metrics}\label{sec:performance_metrics}
We quantify the added value of rationale information for our two learning paradigms. Both $TC$ and $AR$ are ratios indicating the performance over a baseline. Compared to absolute differences, ratios make relative improvements interpretable and comparisons meaningful across tasks. 

\paragraph{Learnability metric \textit{TC}} We train a binary token classifier $\mathcal{T}$ on rationales. $TC$ is defined as: 
\begin{equation}
    TC = \text{token-f1}_{\mathcal{T}(D)} / \text{token-f1}_{\mathcal{B}(D)}
\end{equation}
\noindent It measures the relative performance of the token classifier $\mathcal{T}$ on rationales as token labels (computed at the instance-level and averaged over dataset $D$) compared to a dataset-specific baseline function $\mathcal{B}$ that assigns the majority label to each token. We use this simple baseline as a proxy to contextualise model performance within task-specific complexity (not for benchmarking purposes) making cross-task comparisons more meaningful. A relative score also accounts for different baseline $\mathcal{B}$ performance across models since token labels differ due to different tokenisations. The greater $TC$, for $TC$$\scriptstyle >$$1$, the greater the improvement. We expect a high $TC$ for informative rationales, i.e., for rationales where $CI$ is low.

\paragraph{Learnability metric \textit{AR}} We fine-tune a sequence classifier on inputs that are rationalised during training only, through attention regularisation. $AR$ is computed as: 
\begin{equation}
    AR = \text{f1}_{\mathcal{R}(D)} / \text{f1}_{\mathcal{M}(D)}
\end{equation}
\noindent It represents the relative performance of the regularised model $\mathcal{R}$ with respect to the baseline model $\mathcal{M}$ on the non-rationalised test set. The higher $AR$, for $AR$$\scriptstyle >$$1$, the more the rationale regularization improved the model $\mathcal{M}$. 

The input of $\mathcal{R}$ during training is a tokenised sentence with a sentence-wise label $y$ and a binary rationale mask $a$ (1 for important tokens, else 0) serving as ground truth for guiding the attention mechanism. During the forward pass, we extract the attention weights $\hat{a}$. The regularised model is then optimised with a second loss term $\mathcal{L}_{\text{attention}}$ (binary cross-entropy) measuring the distance between rationale mask $a$ and attention weights $\hat{a}$:

\begin{figure*}[htbp]
    \centering
    \includegraphics[width=1.0\textwidth]{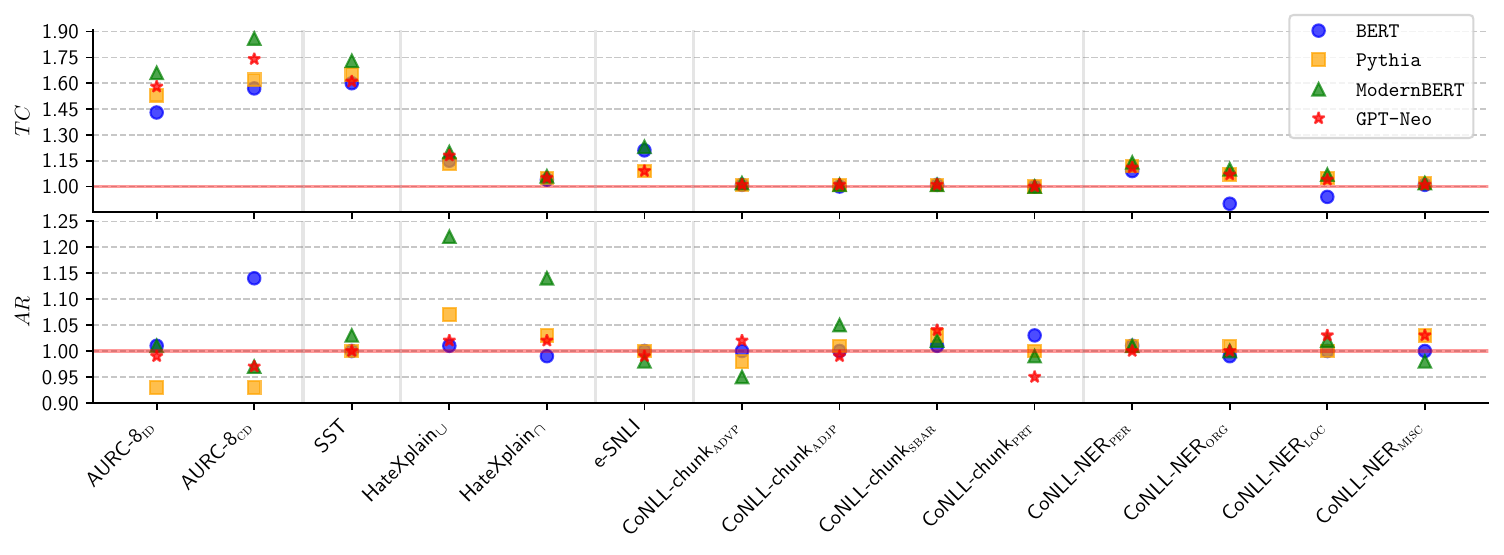}
    \caption{Performance results expressed in $TC$ and $AR$. Scores $> 1.00$ indicate model improvement over the baselines described in §\ref{sec:performance_metrics}. For example, $1.15$ indicates a relative improvement by a factor of $1.15$, or +15\%.}
    \label{fig:raw_performance}
\end{figure*}

\begin{equation}
    \mathcal{L} = \mathcal{L}_{\text{task}}(y, \hat{y}) + \mathcal{L}_{\text{attention}}(a, \hat{a})
\end{equation}

\noindent \citet{kozlova2024transformer} explored the attention weights from the first, last and a specific layer for which human eye-tracking attention and model correlated most. Due to the extensive fine-tuning, we confine our experiments to the attention values of the last layer as it is closest to the final predictions. 


When we isolate the effect of attention regularisation with rationales, we aim to understand the rationales' informativeness in guiding the model and assess whether regularisation is a useful technique. If we do not observe performance improvements, this might be due to the rationale being redundant (the model would already rely on similar cues) or detrimental (steering the model away from dataset-specific shortcuts that boosted original results). Attention regularisation might optimise better for \texttt{BERT} and \texttt{ModernBERT} (than for \texttt{Pythia} and \texttt{GPT-Neo}) because of the ability to attend to all tokens rather than only the past tokens in the input.

\subsection{Results of the Two Learning Paradigms}\label{sec:observed_diff_mod_perf}
We describe the effects of attention regularisation through $AR$ and the ability of classifying rationale tokens through $TC$. We summarise three main observations based on Figure \ref{fig:raw_performance}. More details are provided in Appendix \ref{appendix:complementary_overview_tables}, including the absolute f1 scores of the models and baselines (Figure~\ref{fig:f1_improvements}, Table~\ref{tab:overview}) and results for significance tests (Table~\ref{tab:AR_confintervals}). 

\paragraph{Regularisation is task- and model-dependent} The $AR$ results in Figure \ref{fig:raw_performance} show that attention regularisation has a three-run-average positive effect in 6/14 tasks for \texttt{BERT}, 7 for \texttt{Pythia}, 8 for \texttt{ModernBERT}, and 6 for \texttt{GPT-Neo}. To determine the consistency of observed improvements, we compute 95\% bootstrap confidence intervals by extracting the averages of three samples with replacement (1,000 iterations) from the performances of the regularised and baseline models, for each model$\times$task setup. Given the small sample size of 3 runs per setup, these confidence intervals must be interpreted with caution. We find that the lower bounds of the intervals indicate consistent improvement in 83\% of cases for \texttt{BERT}, even for relatively low $AR$, e.g.\ $<1.05$. Consistent improvements are lower for the other three models: 43\%, 25\% and~0\%. The latter (\texttt{GPT-Neo}) can be explained by relatively low $AR$ overall, peaking at $1.04$. These partly divergent results show that the effectivity of attention regularisation is, besides the expected differences among tasks, highly model dependent. We also find that regularisation does not guarantee a greater stability for predictions on the same instances between models, contrary to the expectation that rationalisation would mitigate the differences between random seeds (Appendix \ref{appendix:secondary_results}).

\paragraph{Rationales can boost X-domain performance}
We take a closer look at in- vs. cross-domain argument mining (AURC-8\textsubscript{\textsc{id}}, AURC-8\textsubscript{\textsc{cd}}). We partly validate the idea that attention regularisation improves performance in both in- and cross-domain settings \citep{hartmann-sonntag-2022-survey}. In this specific task, only \texttt{BERT} benefits from attention regularisation on both settings, with cross-domain performance gaining a substantially greater improvement than in-domain ($AR = 1.14$ vs. $AR = 1.01$); the former problem being inherently more difficult appears to be compensated by explicit guidance. 
Notably, by enhancing sequence classifiers (comparable to the ones by \citet{trautmann2020fine}) with a simple loss term and the available rationale data that they use for token classification, we drastically reduce the gap between cross-domain and in-domain argument mining. 
At the same time, \texttt{ModernBERT} only improves on average but non-consistently in the in-domain setting ($AR = 1.01$, [$0.99$, $1.02$]), while the two autoregressive models do not benefit from rationale-based regularisation. 

\paragraph{\texttt{ModernBERT} $>$ autoregressive models on \textit{TC}} 
We expected that our masked language models would perform best on token classification. While $TC$ is higher for \texttt{Pythia} and \texttt{GPT-Neo} on 9 out of 14 tasks compared to \texttt{BERT} (Figure \ref{fig:raw_performance}), \texttt{ModernBERT} outperforms the former two on at least 10 tasks. Still, no strong model generalisations can be drawn from this small sample size. Interestingly, the classifiers with the highest $TC$ are the ones that were trained on the datasets with highest rationale density $|r|/|x|$. This is likely because the majority class baseline is higher for tasks where the rationales are less dense (Table \ref{tab:overview}, Appendix \ref{appendix:complementary_overview_tables}).

\subsection{Relating Dataset-\textit{CI} to Learnability}\label{sec:relating_suff_perf}
The relation between $CI$ and learning from rationales remains unclear: what is it that $CI$ captures about rationales that a model can use? Hence, the question we address is whether $CI$ aligns with the learnability metrics $TC$ and $AR$. 

As a first exploratory step, we analyse the relation between dataset-average $CI$ (Figure \ref{fig:raw_sufficiency}) and performance. We therefore compute pairwise correlations on $(CI, TC)$, $(CI, AR)$. The learnability metrics ($\frac{\text{f1}}{\text{bl}}$, where \textit{bl} is the baseline) are first normalised as $(\frac{\text{f1}}{\text{bl}} \cdot \text{bl}) - \text{bl}) / {(1 - \text{bl})}$ to reflect the relative improvement to both the baseline and the upper performance limit ($1.0$ f1), giving a more balanced weight to the improvement of models that have a high base performance. Given the small sample size (6 condensed task versions to ensure statistical independence) the correlation's magnitude may be unstable but can still provide indications of polarity. We adopt Spearman's $r$ because of its higher outlier robustness compared to e.g.\ Pearson's $r$. We hypothesise negative correlations between dataset-level $CI$ and performance based on the intuition that low $CI$ entails high rationale informativeness, hence stronger cues for guiding the model. 

\paragraph{\textit{CI} might correlate with \textit{AR} for \texttt{BERT}} 
Cautiously in line with our expectations, we find negative correlations that are moderate for $AR$ with \texttt{BERT}, yet results are not significant ($r = -.49$, $p = .33$). We observe non-significant absent to weak correlations for the other models and no correlation between $CI$ and $TC$. As these signs are inconclusive, we extend this small dataset-level analysis with a full instance-level analysis (§\ref{sec:instance-level-analysis}), aiming to solidify the understanding of $CI$ and its ties to model performance.

\subsection{Relating Instance-\textit{CI} to Learnability}\label{sec:instance-level-analysis}
If low $CI$ really entails high rationale informativeness, we would expect $CI$ to correlate (again, negatively) with prediction correctness of the inputs. Furthermore, we expect that instances with low $CI$ either tend to receive (i) correct predictions by base model $\mathcal{M}$ or (ii) initial incorrect predictions by $\mathcal{M}$ that are flipped by the regularised model $\mathcal{R}$. We investigate this by analysing the pairs of tasks with respectively the highest and lowest performance increase ($AR$).\footnote{
\uline{\texttt{BERT}}: \mbox{AURC-8\textsubscript{\textsc{cd}}} \& HateXplain\textsubscript{$\cap$};
\uline{\texttt{Pythia}}: HateXplain\textsubscript{$\cup$} \& \mbox{AURC-8\textsubscript{\textsc{id}}};
\uline{\texttt{ModernBERT}}: HateXplain\textsubscript{$\cup$} \& CoNLL-chunk\textsubscript{\textsc{advp}}; \uline{\texttt{GPT-Neo}}: CoNLL-chunk\textsubscript{\textsc{sbar}} \& CoNLL-chunk\textsubscript{\textsc{prt}}.} We expect the relation between $CI$ and predictions to be stronger in tasks for which $\mathcal{R}$ successfully learnt from the rationales (high $AR$) than in tasks where attention regularisation had a non-positive effect (low $AR$). We cannot compare models directly to one another as the selected tasks vary; instead, we inspect them in parallel.

\begin{figure}[htbp]
    \centering
    \includegraphics[width=0.49\textwidth]{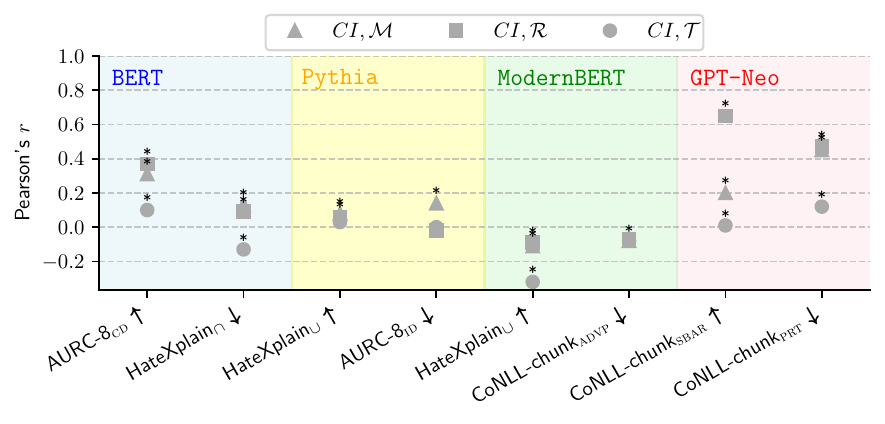}
    \caption{The correlation between $CI$ and predictions in high $AR$ (↑) and low $AR$ (↓) tasks tends to positive, not negative, suggesting that high $CI$ does not entail rationale informativeness. Significance (*) at $p$ < .05.}
    \label{fig:correlation_plot}
\end{figure}

\paragraph{No negative correlation, but sometimes positive}
Figure \ref{fig:correlation_plot} shows Pearson's correlation between instance-level $CI$ (continuous scores) and predictions. Prediction correctness is binary\footnote{In this case, we compute the point-biserial correlation, which is equivalent to Pearson's $r$ for continuous variables.} for the sequence classifiers $\mathcal{M}$ and $\mathcal{R}$; predictions by token classifier $\mathcal{T}$ are continuous f1-scores based on token label correctness. For \texttt{BERT}, contrary to our expectations, the correlation with base model $\mathcal{M}$ and regularised model $\mathcal{R}$ is weak to moderately \textit{positive} for the high $AR$ task, and weak positive to absent for the low $AR$ task. We observe the same drop for $\mathcal{T}$, but the correlation remains absent. For \texttt{Pythia}, we see a different result: $\mathcal{T}$ predictions still do not correlate, but neither do $\mathcal{M}$ and $\mathcal{R}$ (weak to absent, mostly non-significant). While no correlation is detected for \texttt{ModernBERT}, up to strong positive correlations are found for \texttt{GPT-Neo}, but with no clear differences between high and low $AR$ tasks.

\begin{figure*}[h!]
    \centering
    \includegraphics[width=1.0\textwidth]{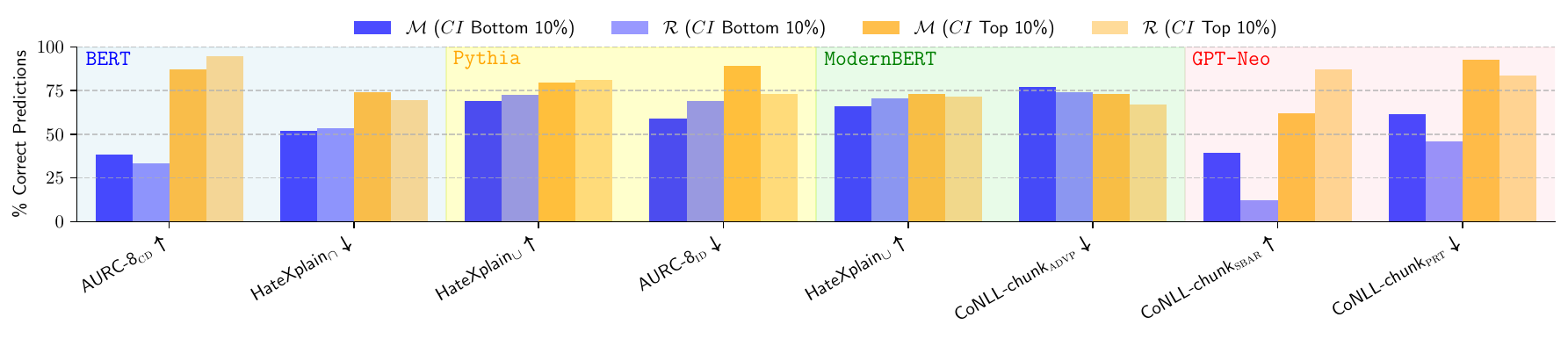}
    \caption{\% Correct predictions for bottom and top Contextual Impact. For each model, we compare the respective tasks that achieved highest $AR$ (↑)--where a positive rationalisation effect was observed, vs. lowest $AR$ (↓). Overall, high $CI$ leans to correct predictions. A greater top--bottom distance for $\mathcal{R}$ is observed in high $AR$ (↑) tasks.}
    \label{fig:topbottomchart_plot}
\end{figure*}

\paragraph{\textit{\textbf{CI}} meaning becomes clearer at the extremes}
Correlations may be mitigated by the most common values (the middle of the $CI$ distribution). In Figure \ref{fig:topbottomchart_plot}, we thus inspect the extremes: instances with bottom and top 10\% $CI$ where the link with predictions should be more pronounced. We find that the top 10\% $CI$ instances are tied to overall better predictions than the bottom 10\%, strengthening the signs of positive correlation (Figure \ref{fig:correlation_plot}). 

This tendency for \textit{high $CI$ implying better predictions} stands in contrast to our initial hypotheses. If we update our hypotheses based on this inverse tendency, we would accordingly expect top $CI$ instances to receive better predictions than bottom $CI$ instances when zooming in on the regularised model $\mathcal{R}$. $\mathcal{R}$ is of interest as it allows us to differentiate between models that did and did not learn from the rationales. Specifically, we expect $\Delta\text{Pred}$, the prediction performance distance between top and bottom $CI$ instances, to be stronger in high $AR$ tasks ($\uparrow$), i.e.\ tasks where rationalisation had a positive effect, compared to low $AR$ tasks ($\downarrow$).
Based on Figure \ref{fig:topbottomchart_plot}, we confirm this expectation in all four scenarios: for \uline{\texttt{BERT}} ($\Delta\text{Pred}\uparrow=62 >\Delta\text{Pred}\downarrow=15$) and \uline{\texttt{GPT-Neo}} ($ 75>38$), $\Delta\text{Pred}$ is clearly stronger in the $\uparrow AR$ task. For \uline{\texttt{Pythia}}, the expected contrast between $\uparrow AR$ and $\downarrow AR$ tasks is present but weaker ($9>4$). For \uline{\texttt{ModernBERT}}, the distance between top and bottom is marginal in the $\uparrow AR$ task ($\Delta\text{Pred}\uparrow=1$), while the prediction performance for $\mathcal{R}$ on top is even below bottom in the $\downarrow AR$ task ($\Delta\text{Pred}\downarrow=-7$). More details about the distances are given in Table \ref{tab:delta-pred}, Appendix \ref{appendix:complementary_overview_tables}.

By comparing the regularised model $\mathcal{R}$ and base model $\mathcal{M}$, we finally find that attention regularisation improves predictions especially on high $CI$ instances for models that successfully learnt from rationales (↑$AR$ tasks), except for \texttt{ModernBERT}. 
 
In summary, the fact that \textit{high} instance $CI$ leans to \textit{correct} predictions seems to indicate that $CI$ does not capture rationale informativeness in the way we originally hypothesised. This finding also counters the weak correlation direction found at the dataset level in §\ref{sec:relating_suff_perf}. 
These instance-level results suggest that \textit{low} $CI$ does not tell much about the absolute informativeness of the rationale (to improve a model). Rather, $CI$ appears to capture information about the way in which rationales and non-rationale contexts interact with one another. At least at the extremes of the $CI$ spectrum, \textit{high} $CI$ captures the impact of the context information on the prediction, i.e.\ in the form of interference with rationale information in the same input. 

\section{\textit{\textbf{CI}} and Rationale Aggregation}\label{sec:sufficiency_and_rationale_aggregation}
Since rationale quality is assumed to be central to model improvement, we included two versions of HateXplain (taking advantage of the multiple annotators per instance) to test the effect of different rationale aggregation strategies on model performance and $CI$. Do rather few, strong rationale words or many, noisy rationale words guide the model better? In theory, for $CI$ to be low, rationales should include key words that are informative about the sentence label and exclude context words that could bias the prediction. Rationale annotation is subject to human disagreement, which arguably defines a rationale's informativeness to some degree: by aggregating rationale annotations through union, these unified rationales would include relatively many words, part of which would have a disputed informativeness ($0$$\scriptstyle <$annotators$\scriptstyle <$$n$ highlighted the word). In contrast, rationales aggregated by intersection would strictly include key words (where all $n$ annotators agreed on), but lack those disputed words that might carry partial information about the target class. Figure \ref{fig:raw_sufficiency} shows that $CI$ is lower for HateXplain\textsubscript{$\cup$} than for --\textsubscript{$\cap$} (for all four models), with also $TC$ and $AR$ being mostly higher (cautiously in line with our negative correlation hypothesis from §\ref{sec:relating_suff_perf}).

This suggests that the incorporation of disputed rationales in the input benefits modelling, at least in this scenario of hate speech detection. In other words, collecting different human perspectives on word importance (aggregated through union) is preferred over a single annotator that might label too strictly. In cases where a single annotator per instance is available, the degree to which they act as a \textit{lumper} is thus likely to affect the quality of the collected rationales. In combination with the granularity restrictions given in the annotation instructions, this annotator characteristic potentially influences $CI$ and the learning ability of the model.  

\section{Conclusion}
We carried out a rich set of experiments investigating the role of sufficiency, reframed as $CI$, in capturing rationale informativeness for model improvement on two different learning paradigms. What is important information and is there a link in the way models process it? Well, it depends: \textbf{(A)} There are signs, although thin, that low dataset-level $CI$ encodes high rationale informativeness for regularised \texttt{BERT} models, cautiously in line with our initial hypotheses. \textbf{(B)} However, our instance-level analyses yield a pattern that on the one hand is clearer but also goes against expectations, pointing to \textit{higher CI, better predictions} (visible at the extremes of the $CI$ spectrum). This suggests that high $CI$ inputs rely on the informativeness stemming from \textit{both} rationales and non-rationale contexts. In other words, high $CI$ would entail the context words' \textit{relative weakness} (in interfering with rationale information in the same input), rather than reflecting in any way the rationales' \textit{absolute strength}. To this end, \textbf{\textit{contextual impact}} appropriately renames sufficiency. Furthermore, \textbf{(C)} there is no one-size-fits-all solution on how to rationalise a model because information value is strongly determined by the task and by the processor (the model using the rationale or not, the humans disagreeing on its length or content). These heterogeneous results show that, overall, sufficiency is not a good indicator of how to prioritise information in training a model. Within this heterogeneity, though, \textbf{(D)} attention regularisation using rationales shows potential as it closes the gap between in- and cross-domain argument mining performance for \texttt{BERT} by adopting a simple auxiliary loss. \textbf{(E)} Finally, the learnability of a token classifier does not correlate with $CI$. This surprising finding shows that the relation between rationales and their context is complex and sufficiency can mask underlying processes, e.g.\ models that perform well on test data due to shortcuts in the context can have high $CI$ despite informative rationales.

Overall, results show that the informativeness of rationales merits further investigation. We would like to extend our analyses to more datasets including other languages (following \citet{kozlova2024transformer} on Russian). Investigating datasets in different languages has the additional advantage that it can provide insight into both language- and culture-specific components in human perception of sufficient information, which especially arises in unconstrained rationales. Finally, new insights about contextual information can lead to reconsidering \textit{comprehensiveness} (§\ref{sec:faithfulness_assessment}), i.e.\ faithfulness through rationale ablation, to further investigate the balance between rationale and context information.



\section*{Limitations}
As different models are pre-trained using different tokenisers, the alignment between (sub)tokens and rationales may also differ. When comparing models that are pre-trained differently, there is no straightforward solution. Secondly, default hyperparameters were used with only slight adaptations. Keeping hyperparameters mostly constant (i) reduces their confounding impact and (ii) diminishes the ecological footprint and costs. With more extensive hyperparameter tuning, we would achieve more optimally fine-tuned models for the different tasks and datasets. Although we remained cautious in our claims, there always remains a risk that factors such as the above interfere with the interpretation of our model comparisons.

\section*{Acknowledgements}
This research is funded by the Dutch National Science Organisation (NWO) through project code NWA.1292.19.399 -- \textit{InDeep: Interpreting Deep Learning Models for Text and Sound}. Lisa Beinborn' research is supported by an \emph{Impulsprofessur} grant from the \emph{zukunft.niedersachsen} program and by a VENI grant (Vl.Veni.211C.039) from the Dutch National Science Organisation (NWO). We thank the anonymous reviewers for their constructive comments that helped improve the paper. Remaining errors are our own.


\bibliography{custom}

\appendix
\section{Appendix}\label{appendix:A}
This appendix includes a set of secondary results (\ref{appendix:secondary_results}), a report of the main technical details of our experiments (\ref{appendix:technical_details}) and a collection of complementary overview visualisations (\ref{appendix:complementary_overview_tables}).

\subsection{Secondary Results}\label{appendix:secondary_results}

\begin{figure*}[htbp]
    \centering
    \includegraphics[width=1.0\textwidth]{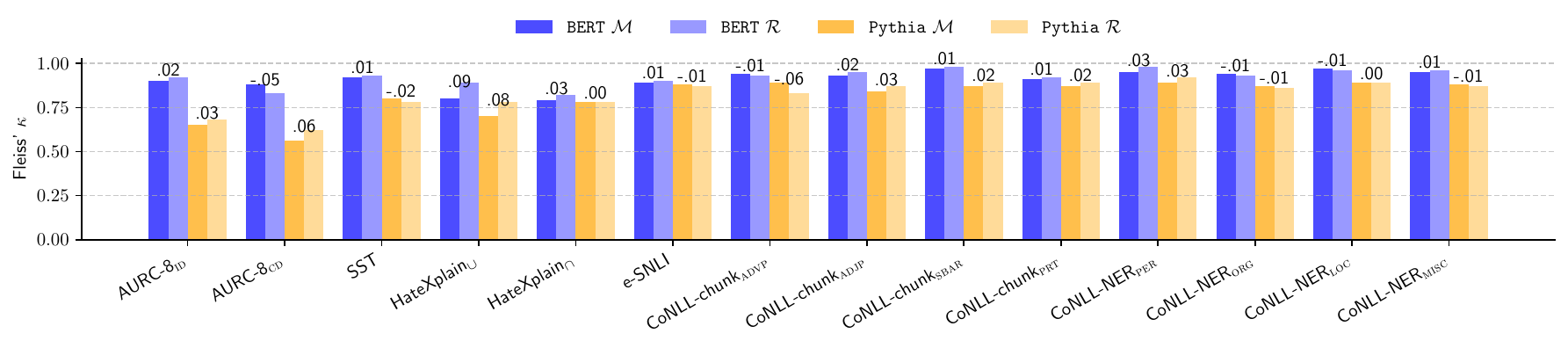}
    \caption{Inter-model agreement (Fleiss' $\kappa$) on predictions given by three different runs. The agreement differences between base model $\mathcal{M}$ and regularised model $\mathcal{R}$ are reported on top of each of the model--task combinations.}
    \label{fig:fleiss}
\end{figure*}

\paragraph{\texttt{BERT}'s $\mathcal{M}$ and $\mathcal{R}$ are more stable than \texttt{Pythia}} By explicitly guiding the model to use rationales for a prediction, we expected that the agreement between models fine-tuned on different random seeds would stabilise. Through Fleiss' $\kappa$ \citep{fleiss1971measuring}, we measure the inter-model agreement between class label predictions on the test set, from three differently seeded runs. Figure \ref{fig:fleiss} shows that \texttt{BERT} predictions are more stable both on the base model $\mathcal{M}$ and on the model with attention regularisation $\mathcal{R}$. Besides the overall higher stability, \texttt{BERT} also exhibits more agreement increases (10 vs. 8) and the strongest improvement ($+.09$ on HateXplain\textsubscript{\textsc{$\cup$}}). \texttt{Pythia} has the biggest drop ($-.06$ on CoNLL-chunk\textsubscript{\textsc{advp}}).

\begin{figure}[htbp]
    \centering
    \includegraphics[width=0.49\textwidth]{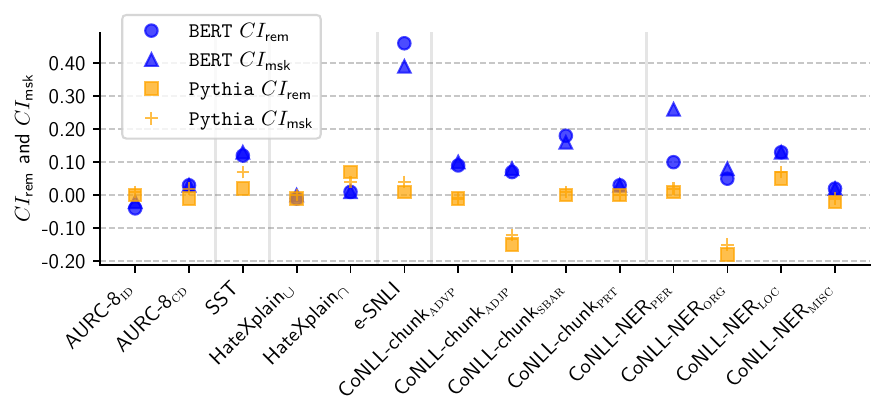}
    \caption{Different implementations of Contextual Impact compared: $CI_\text{rem}$ and $CI_\text{msk}$ are similar.}
    \label{fig:raw_sufficiency_full}
\end{figure}

\paragraph{$CI_{\text{rem}}$ and $CI_{\text{msk}}$ strongly correlate for \texttt{BERT}}
Prior assumptions stated that different implementations of $CI$, i.e.\ by removing ($CI_{\text{rem}}$) or by masking ($CI_{\text{msk}}$) the non-rationales from the input, may yield contrasting results. Figure \ref{fig:raw_sufficiency_full} shows that different implementations of $CI$ are similar for \texttt{BERT} and \texttt{Pythia}. We compare the absolute Contextual Impact scores (correlation and polarity) in the light of a model comparison between \texttt{BERT} and \texttt{Pythia}. We expect the differences to be smaller for \texttt{BERT} as it should recognise masked and removed tokens in a similar way due to its masked language modelling in pretraining. We find that the correlation between the two is positive strong for both models (full overview in Table \ref{tab:corr_combined}. Only Kendall's correlation is significant for \texttt{BERT} ($\tau = .87$, $p < .05$), which is however notable given the small sample. We find that 13 out of 14 task versions have same polarity for \texttt{BERT}, and 12 for \texttt{Pythia}. The slightly higher polarity error and the non-significant correlation by \texttt{Pythia} can be explained by the lack of an associated meaning of the [MASK] token for not being pre-trained as a masked language model. However, the difference is minimal, indicating that it may still be appropriate to compute $CI$ through removal or masking interchangeably for left-to-right models.

\begin{table}[htb]
\small
\centering
\begin{tabularx}{\columnwidth}{l *{4}{>{\centering\arraybackslash}X}}
\toprule
& \multicolumn{2}{c}{\textbf{\texttt{BERT}}} & \multicolumn{2}{c}{\textbf{\texttt{Pythia}}} \\
\cmidrule(lr){2-3} \cmidrule(lr){4-5}
\textbf{Variables} & \textbf{$r$} & \textbf{$\tau$} & \textbf{$r$} & \textbf{$\tau$} \\
\midrule
$CI_{\text{rem}}, CI_{\text{msk}}$ & .94 & \textbf{.87*} & .83 & .73 \\
$CI_{\text{rem}}, TC$              & .14 & .07         & -.09 & .07 \\
$CI_{\text{rem}}, AR$              & -.49 & -.33       & .14 & .20 \\
$CI_{\text{msk}}, TC$              & .09 & -.07        & .09 & .07 \\
$CI_{\text{msk}}, AR$              & -.54 & -.47       & -.26 & -.07 \\
\bottomrule
\end{tabularx}
\caption{Pairwise correlations for \texttt{BERT} and \texttt{Pythia}. The significance (*) threshold is set at $p<.05$.}
\label{tab:corr_combined}
\end{table}

\paragraph{$CI$ diverges with $\mathcal{R}$} We compare the average, absolute distance from 0 for $CI_{\text{rem}}$ and $CI_{\text{msk}}$. The differences between the two metrics are negligible for both models. Standard deviations are larger for \texttt{BERT} ($.11$ and $.11$ versus $.05$ and $.04$). Until this moment, we measured $CI$ for the baseline model $\mathcal{M}$, but we measure it for the regularised model $\mathcal{R}$, too. If $\mathcal{R}$ learnt from the rationales and $CI$ is reflecting this, the difference in probabilities on the contrast examples (i.e. $CI$) should be closer to 0 than it was for $M$. Table \ref{tab:suff_magnitude_averages} shows that this is not the case: $CI$ scores for $\mathcal{R}$ are instead more distant and/or exhibit a greater standard deviation. 

\paragraph{Polarity does not imply successful modelling} We explored whether negative $CI$ relates to successful $TC$ training (>1), which resulted to be true in a smaller percentage of the cases. Similarly, we find no pattern in the inverse relation between $CI$ polarity and $AR$ success, but this hypothesis is more likely to be true for \texttt{Pythia} (see Table \ref{tab:polarity}). 

\paragraph{\texttt{BERT} displays 'positive' behaviour} In Figure \ref{fig:raw_sufficiency}, we see that $CI$ for \texttt{BERT} is mostly higher and more often positive than \texttt{Pythia}. This may indicate that, for \texttt{BERT}, uncontextualised rationales are less informative than for \texttt{Pythia}.

\begin{figure*}[htbp]
    \centering
    \includegraphics[width=1.0\textwidth]{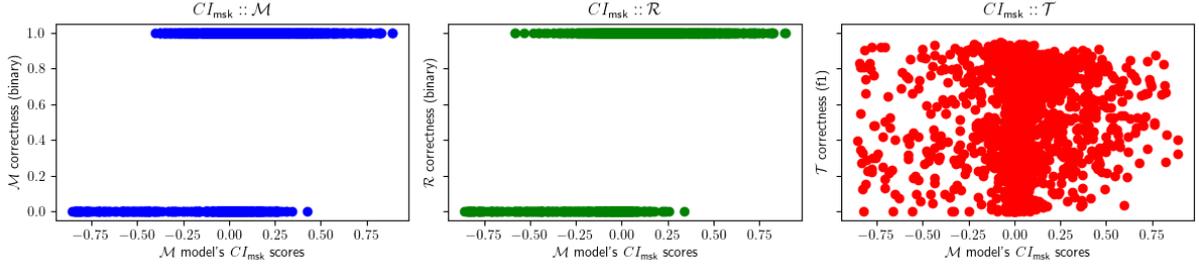}
    \caption{Sufficiency scores in relation to prediction correctness. In this example: $CI_{\text{msk}}$, \texttt{BERT}, AURC-8\textsubscript{{\textsc{cd}}}.}
    \label{fig:scatterplot_suff_pred_corr}
\end{figure*}

\paragraph{Learning from rationales entails $CI$ thresholds} The results in Figure \ref{fig:correlation_plot} and Figure \ref{fig:topbottomchart_plot} are reflected in Figure \ref{fig:scatterplot_suff_pred_corr}. Here, we see that for \texttt{BERT}--AURC-8\textsubscript{cd} (the model that learned most from attention regularisation), for all models $\mathcal{M}$, $\mathcal{R}$ and $\mathcal{T}$, sufficiency is mostly cluttered around 0, but that the extremes show a different behaviour. While there is no indication of correlation for $\mathcal{T}$, $\mathcal{M}$ and $\mathcal{R}$ show a clear threshold: below $-0.50$ the predictions are incorrect, above $0.50$ the predictions are correct. 

\begin{table}[htbp]
\small
\centering
\begin{tabularx}{\columnwidth}{l *{4}{>{\centering\arraybackslash}X}}
    \toprule
    & \multicolumn{2}{c}{\textbf{\texttt{BERT}}} & \multicolumn{2}{c}{\textbf{\texttt{Pythia}}} \\
    \cmidrule(lr){2-5}
    \textbf{Model} & $CI_\text{rem}$ & $CI_\text{msk}$ & $CI_\text{rem}$ & $CI_\text{msk}$\\ 
    \midrule
    $\mathcal{M}$ & .10\textsubscript{$\pm$.11} & .10\textsubscript{$\pm$.11} & .04\textsubscript{$\pm$.05} & .04\textsubscript{$\pm$.04} \\ 
    $\mathcal{R}$ & .10\textsubscript{$\pm$.11} & .11\textsubscript{$\pm$.13} & .05\textsubscript{$\pm$.08} & .04\textsubscript{$\pm$.06} \\  
    \bottomrule
\end{tabularx}
\caption{Average sufficiency scores for $\mathcal{M}$ and $\mathcal{R}$.}
\label{tab:suff_magnitude_averages}
\end{table}

\begin{table}[htbp]
\small
\centering
\begin{tabularx}{\columnwidth}{l *{4}{>{\centering\arraybackslash}X}}
    \toprule
    & \multicolumn{2}{c}{\textbf{\texttt{BERT}}} & \multicolumn{2}{c}{\textbf{\texttt{Pythia}}} \\
    \cmidrule(lr){2-5}
    \textbf{Metric} & $CI_\text{rem}$ & $CI_\text{msk}$ & $CI_\text{rem}$ & $CI_\text{msk}$ \\ 
    \midrule
    $TC$ & 29\% & 21\% & 50\% & 36\% \\ 
    $AR$ & 29\% & 21\% & 36\% & 36\% \\ 
    \bottomrule
\end{tabularx}
\caption{Tasks where polarity is a performance indicator.}
\label{tab:polarity}
\end{table}

\subsection{Technical Details}\label{appendix:technical_details}
\paragraph{Special tokens} Sentences in e-SNLI are concatenated with <sep> token in \texttt{BERT} setup, but no separator token is used for \texttt{Pythia} (eos\_token would not take into consideration the context left of the token). [MASK], [UNK] and [PAD] tokens are not present in \texttt{Pythia} tokenizer. Knowing that they are interpreted differently than in the \texttt{BERT} setup, but to avoid subtokenization (e.g. into "[", "MASK", "]"), we add them to the pre-assigned placeholders in the \texttt{Pythia} tokenizer \texttt{(mask|unk|pad)\_token} as random embeddings.

\paragraph{Rationale--token alignment}
Humans annotate rationales at the word level. When splitting words into (sub)tokens, we need to re-align the rationales. Before tokenising, we assign a rationale label at the character-level. After tokenising, we ignore the subtoken indicators "\#\#" and "Ġ" and remove non-ascii characters for which BPE tokenisation adds artifact characters to the tokenised strings, interfering with alignment. We then collapse character-level rationales to token-level rationales.

\paragraph{Software and hparams}\label{appendix:software_and_hyperparams} 
For each \texttt{<name>} among \texttt{(Bert|GPTNeoX|ModernBert|GPTNeo)}
we use \texttt{<name>ForTokenClassification} and a custom \texttt{<name>Model} from Huggingface's Transformers library, v4.44.2 \citep{wolf2019huggingface}. We fine-tune for up to 10 epochs with a batch size of 16 and keep the checkpoint with the lowest evaluation loss. We kept the learning rate for sequence classification at 3{e}--5 with some exceptions: 1{e}--5 for \texttt{GPTNeo} and 3{e}--6 for some of the \texttt{ModernBERT} runs. 3{e}--5 was used for token classification. A learning rate decay of 1e--2 was used in all experiments. More details are found in our repository.

\paragraph{Hardware and runtimes}\label{appendix:runtimes_and_hardware} Table \ref{tab:runtimes}: models are fine-tuned on either NVIDIA A100-SXM4-40GB (40960 MiB) or on NVIDIA GeForce RTX 2080 Ti (11264 MiB), the choice of which does not affect runtime of the same setup with different random seeds (relatively small models and constant training batch sizes for comparability reasons). Rationalisation of the input does not affect runtime, nor does the dataset version. Token classifier runtimes are comparable.

\begin{table}[h!]
\small
\centering
\begin{tabularx}{\columnwidth}{l *{1}{>{\raggedleft\arraybackslash}X}}
\toprule
\multicolumn{1}{l}{\textbf{Dataset}} & \multicolumn{1}{r}{\textbf{Runtime}} \\
\midrule
AURC-8        & 3--4 min \\
SST           & 4--6 min \\
HateXplain    & 17--25 min \\
e-SNLI        & 7--8 h \\
CoNLL-chunk    & 5--8 min \\
CoNLL-NER    & 13--19 min \\
\bottomrule
\end{tabularx}
\caption{Approximate runtime of a single model fine-tuning on sequence classification tasks for 5 epochs.}
\label{tab:runtimes}
\end{table}

\paragraph{Rationale derivation algorithm for SST}\label{appendix:rationale_derivation_algorithm_SST}
We follow \citet{carton-etal-2020-evaluating} to derive the binary token-level rationale mask from the raw SST dataset. We report their description for completeness in the present appendix. Specifically, SST contains syntactic parse trees of movie reviews, where each node (i.e. phrase) is annotated with a sentiment label ranging from -2 (very negative) to +2 (very positive). Starting with the root and traversing the tree breadth-first, we include a node in the rationale (i.e. we assign a value of 1 to each of the node's tokens) if the sentiment score of the node is greater than all of its constituents. In this way, only the smallest constituents that explain the sentiment are added to the rationale, maximising the granularity of the rationale.

\subsection{Complementary Overview Visualisations}\label{appendix:complementary_overview_tables}
\begin{figure*}[htb]
    \centering
    \includegraphics[width=1.0\textwidth]{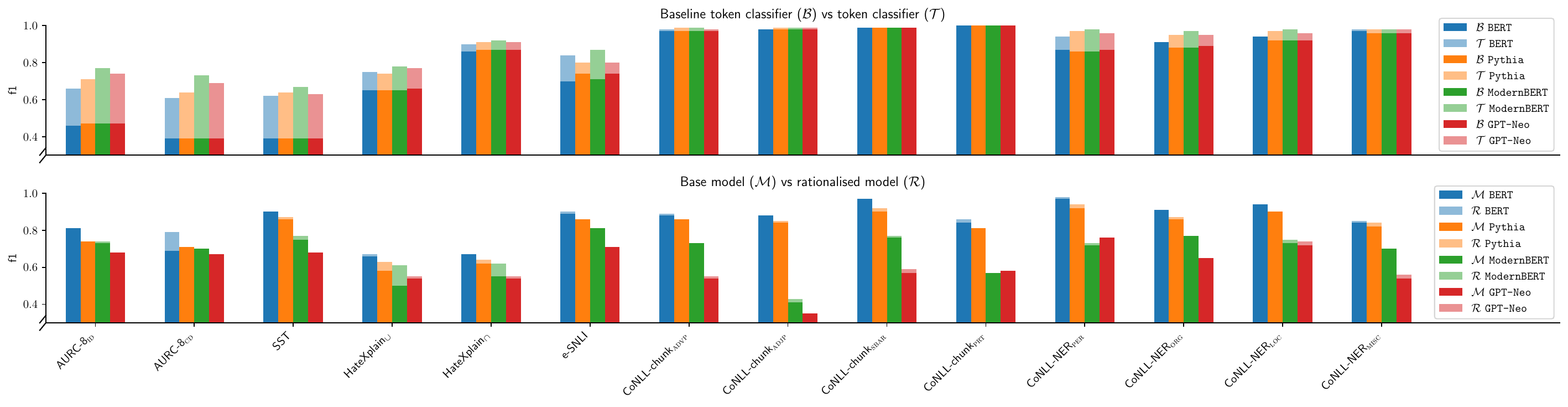}
    \caption{Performance improvements (f1) of the classifiers trained with rationales compared to the baselines.}
    \label{fig:f1_improvements}
\end{figure*}

Figure \ref{fig:f1_improvements} visualises the performance improvements of the classifiers along with their baselines. We then share the full learnability metrics' scores in Table \ref{tab:overview}. Table \ref{tab:AR_confintervals}, instead, represents the full overview of the confidence intervals for $AR$ performance. Finally, Table \ref{tab:delta-pred} contains the computations of $\Delta\text{Pred}$ and accompanies Figure \ref{fig:topbottomchart_plot} in §\ref{sec:instance-level-analysis}. 

\begin{table*}[h!]
\centering
\begin{tabular}{lcc|cc|cc|cc}
\toprule
\multicolumn{1}{c}{\textbf{}} 
& \multicolumn{2}{c}{\textbf{\texttt{BERT}}} 
& \multicolumn{2}{c}{\textbf{\texttt{Pythia}}}
& \multicolumn{2}{c}{\textbf{\texttt{ModernBERT}}}
& \multicolumn{2}{c}{\textbf{\texttt{GPT-Neo}}}\\
\cmidrule(lr){2-3} \cmidrule(lr){4-5} \cmidrule(lr){6-7} \cmidrule(lr){8-9}
\textbf{Task versions} 
& $TC$ & $AR$ 
& $TC$ & $AR$
& $TC$ & $AR$
& $TC$ & $AR$\\
\midrule
AURC-8\textsubscript{\textsc{id}} & 1.43\textsubscript{.463} & 1.01\textsubscript{.808} & 1.53\textsubscript{.467} & 0.93\textsubscript{.741} & 1.66\textsubscript{.467} & 1.01\textsubscript{.733} & 1.58\textsubscript{.467} & 0.99\textsubscript{.677}\\
AURC-8\textsubscript{\textsc{cd}} & 1.57\textsubscript{.391} & 1.14\textsubscript{.692} & 1.62\textsubscript{.393} & 0.93\textsubscript{.712} & 1.86\textsubscript{.393} & 0.97\textsubscript{.702} & 1.74\textsubscript{.393} & 0.97\textsubscript{.674}\\
\midrule
SST & 1.60\textsubscript{.388} & 1.00\textsubscript{.898} & 1.65\textsubscript{.388} & 1.00\textsubscript{.864} & 1.73\textsubscript{.388} & 1.03\textsubscript{.748} & 1.61\textsubscript{.393} & 1.00\textsubscript{.676}\\
\midrule
HateXplain\textsubscript{$\cup$} & 1.15\textsubscript{.648} & 1.01\textsubscript{.660} & 1.13\textsubscript{.652} & 1.07\textsubscript{.582} & 1.20\textsubscript{.652} & 1.22\textsubscript{.499} & 1.18\textsubscript{.655} & 1.02\textsubscript{.542}\\
HateXplain\textsubscript{$\cap$} & 1.04\textsubscript{.863} & 0.99\textsubscript{.673} & 1.05\textsubscript{.866} & 1.03\textsubscript{.621} & 1.06\textsubscript{.866} & 1.14\textsubscript{.547} & 1.05\textsubscript{.869} & 1.02\textsubscript{.541}\\
\midrule
e-SNLI & 1.21\textsubscript{.698} & 1.00\textsubscript{.894} & 1.09\textsubscript{.736} & 1.00\textsubscript{.860} & 1.23\textsubscript{.707} & 0.98\textsubscript{.807} & 1.09\textsubscript{.737} & 0.99\textsubscript{.706}\\
\midrule
CoNLL-chunk\textsubscript{\textsc{advp}} & 1.01\textsubscript{.974} & 1.00\textsubscript{.885} & 1.01\textsubscript{.974} & 0.98\textsubscript{.859} & 1.02\textsubscript{.974} & 0.95\textsubscript{.731} & 1.01\textsubscript{.973} & 1.02\textsubscript{.539}\\
CoNLL-chunk\textsubscript{\textsc{adjp}} & 1.00\textsubscript{.981} & 1.00\textsubscript{.883} & 1.01\textsubscript{.981} & 1.01\textsubscript{.840} & 1.01\textsubscript{.981} & 1.05\textsubscript{.412} & 1.01\textsubscript{.982} & 0.99\textsubscript{.352}\\
CoNLL-chunk\textsubscript{\textsc{sbar}} & 1.01\textsubscript{.986} & 1.01\textsubscript{.967} & 1.01\textsubscript{.986} & 1.03\textsubscript{.895} & 1.01\textsubscript{.986} & 1.02\textsubscript{.755} & 1.01\textsubscript{.985} & 1.04\textsubscript{.570}\\
CoNLL-chunk\textsubscript{\textsc{prt}} & 1.00\textsubscript{.997} & 1.03\textsubscript{.838} & 1.00\textsubscript{.997} & 1.00\textsubscript{.810} & 1.00\textsubscript{.997} & 0.99\textsubscript{.569} & 1.00\textsubscript{.997} & 0.95\textsubscript{.577}\\
\midrule
CoNLL-NER\textsubscript{\textsc{per}} & 1.09\textsubscript{.867} & 1.01\textsubscript{.967} & 1.12\textsubscript{.862} & 1.01\textsubscript{.924} & 1.14\textsubscript{.862} & 1.01\textsubscript{.722} & 1.11\textsubscript{.866} & 1.00\textsubscript{.755}\\
CoNLL-NER\textsubscript{\textsc{org}} & 0.90\textsubscript{.907} & 0.99\textsubscript{.911} & 1.07\textsubscript{.882} & 1.01\textsubscript{.860} & 1.10\textsubscript{.882} & 1.00\textsubscript{.766} & 1.07\textsubscript{.886} & 1.00\textsubscript{.652}\\
CoNLL-NER\textsubscript{\textsc{loc}} & 0.94\textsubscript{.938} & 1.00\textsubscript{.935} & 1.05\textsubscript{.921} & 1.00\textsubscript{.900} & 1.07\textsubscript{.921} & 1.02\textsubscript{.734} & 1.04\textsubscript{.923} & 1.03\textsubscript{.723}\\
CoNLL-NER\textsubscript{\textsc{misc}} & 1.01\textsubscript{.968} & 1.00\textsubscript{.844} & 1.02\textsubscript{.962} & 1.03\textsubscript{.817} & 1.02\textsubscript{.962} & 0.98\textsubscript{.702} & 1.01\textsubscript{.964} & 1.03\textsubscript{.539}\\
\bottomrule
\end{tabular}
\caption{Overview of learnability metrics. $TC$ and $AR$ are both ratios and equal an f1-score divided by a baseline; they are presented with their respective baselines, which are given in subscript. Example: the AURC-8\textsubscript{\textsc{id}} regularised \texttt{BERT} model $\mathcal{R}$ outperforms the baseline model $\mathcal{M}$ (f1 = .808) by 1.01 times (a 1\% increase).}
\label{tab:overview}
\end{table*}

\begin{table*}[ht]
\centering
\begin{tabular}{lcccc}
\toprule
\textbf{Task versions} & \textbf{\texttt{BERT}} & \textbf{\texttt{Pythia}} & \textbf{\texttt{ModernBERT}} & \textbf{\texttt{GPT-Neo}} \\
\midrule
AURC-8\textsubscript{\textsc{id}}  & 1.01 \textsubscript{[1.0022, 1.0090]}  & 0.93 \textsubscript{[0.8467, 0.9883]}  & 1.01 \textsubscript{[0.9926, 1.0213]}  & 0.99 \textsubscript{[0.9713, 1.0133]}  \\
AURC-8\textsubscript{\textsc{cd}}  & 1.14 \textsubscript{[1.1035, 1.1759]} & 0.93 \textsubscript{[0.8265, 1.0454]}  & 0.97 \textsubscript{[0.9339, 1.0167]}  & 0.97 \textsubscript{[0.9113, 1.0145]}  \\
\midrule
SST  & 1.00 \textsubscript{[0.9899, 1.0099]}  & 1.00 \textsubscript{[0.9881, 1.0195]}   & 1.03 \textsubscript{[0.9855, 1.0627]}  & 1.00 \textsubscript{[0.9626, 1.0396]}   \\
\midrule
HateXplain\textsubscript{$\cup$}  & 1.01 \textsubscript{[0.9846, 1.0288]} & 1.07 \textsubscript{[1.0034, 1.1437]}  & 1.22 \textsubscript{[1.1850, 1.2522]}   & 1.02 \textsubscript{[0.9944, 1.0494]}  \\
HateXplain\textsubscript{$\cap$}  & 0.99 \textsubscript{[0.9724, 1.0154]} & 1.03 \textsubscript{[0.9936, 1.0600]}    & 1.14 \textsubscript{[1.0410, 1.2270]}    & 1.02 \textsubscript{[0.9741, 1.0651]}  \\
\midrule
e-SNLI  & 1.00 \textsubscript{[0.9946, 1.0118]}  & 1.00 \textsubscript{[0.9981, 1.0055]}   & 0.98 \textsubscript{[0.9700, 0.9905]}    & 0.99 \textsubscript{[0.9900, 0.9962]}    \\
\midrule
CoNLL-chunk\textsubscript{\textsc{advp}}  & 1.00 \textsubscript{[0.9974, 1.0123]}  & 0.98 \textsubscript{[0.9633, 0.9948]}  & 0.95 \textsubscript{[0.8635, 0.9949]}  & 1.02 \textsubscript{[0.9066, 1.1898]}  \\
CoNLL-chunk\textsubscript{\textsc{adjp}}  & 1.00 \textsubscript{[0.9847, 1.0085]}  & 1.01 \textsubscript{[0.9967, 1.0239]}  & 1.05 \textsubscript{[0.8678, 1.2328]}  & 0.99 \textsubscript{[0.7409, 1.3584]}  \\
CoNLL-chunk\textsubscript{\textsc{sbar}}  & 1.01 \textsubscript{[1.0041, 1.0112]} & 1.03 \textsubscript{[1.0075, 1.0590]}   & 1.02 \textsubscript{[0.9872, 1.0539]}  & 1.04 \textsubscript{[0.9660, 1.1365]}   \\
CoNLL-chunk\textsubscript{\textsc{prt}} & 1.03 \textsubscript{[1.0071, 1.0514]} & 1.00 \textsubscript{[0.9807, 1.0271]}   & 0.99 \textsubscript{[0.8954, 1.1073]}  & 0.95 \textsubscript{[0.8306, 1.0991]}  \\
\midrule
CoNLL-NER\textsubscript{\textsc{per}} & 1.01 \textsubscript{[1.0008, 1.0303]} & 1.01 \textsubscript{[0.9988, 1.0275]}  & 1.01 \textsubscript{[0.9948, 1.0343]}  & 1.00 \textsubscript{[0.9865, 1.0226]}   \\
CoNLL-NER\textsubscript{\textsc{org}} & 0.99 \textsubscript{[0.9864, 0.9960]}  & 1.01 \textsubscript{[0.9913, 1.0251]}  & 1.00 \textsubscript{[1.0011, 1.0095]}   & 1.00 \textsubscript{[0.9661, 1.0477]}   \\
CoNLL-NER\textsubscript{\textsc{loc}} & 1.00 \textsubscript{[0.9969, 1.0034]}  & 1.00 \textsubscript{[0.9956, 1.0106]}   & 1.02 \textsubscript{[0.9512, 1.0783]}  & 1.03 \textsubscript{[0.9953, 1.0567]}  \\
CoNLL-NER\textsubscript{\textsc{misc}} & 1.00 \textsubscript{[0.9945, 1.0138]}  & 1.03 \textsubscript{[1.0055, 1.0548]}  & 0.98 \textsubscript{[0.9616, 1.0056]}  & 1.03 \textsubscript{[0.9995, 1.0650]}   \\
\bottomrule
\end{tabular}
\caption{Mean and 95\% [confidence intervals] for $AR$ performance across models and tasks.}
\label{tab:AR_confintervals}
\end{table*}

\begin{table*}[h!]
\small
\centering
\begin{tabularx}{\textwidth}{l *{6}{>{\centering\arraybackslash}X}}
\toprule
&
\multicolumn{3}{c}{$\uparrow AR$} &
\multicolumn{3}{c}{$\downarrow AR$} \\
\cmidrule(lr){2-4} \cmidrule(lr){5-7}
 \textbf{Model} & Top & Bottom & $\Delta\text{Pred}$ & Top & Bottom & $\Delta\text{Pred}$ \\
\midrule
\texttt{BERT}        & 95 & 33 & \textbf{62}  & 69 & 54 & \textbf{15} \\
\texttt{Pythia}      & 87 & 12 & \textbf{75}  & 84 & 46 & \textbf{38} \\
\texttt{ModernBERT}  & 81 & 72 & \textbf{9}   & 73 & 69 & \textbf{4}  \\
\texttt{GPT-Neo}     & 72 & 71 & \textbf{1}   & 67 & 74 & \textbf{-7} \\
\bottomrule
\end{tabularx}
\caption{Accompanying table for Figure~\ref{fig:topbottomchart_plot}. Comparison of top $AR$ ($\uparrow$) and bottom $AR$ ($\downarrow$) correct predictions for the rationalised model $\mathcal{R}$. We compare their signed differences denoted as $\Delta\text{Pred}$.}
\label{tab:delta-pred}
\end{table*}

\end{document}